\documentclass{article}

\usepackage{amsmath}
\usepackage{amssymb}
\usepackage{mathtools}
\usepackage{amsthm}

\usepackage{multirow}
\usepackage{xcolor}
\usepackage{tcolorbox}
\usepackage{booktabs}
\usepackage{tabularx}
\usepackage{packages/algorithm}
\usepackage{packages/algorithmic}
\usepackage{xurl}   
\usepackage{hyperref}
\usepackage[capitalize,noabbrev]{cleveref}
\hypersetup{breaklinks=true}

\newcommand{\calM}{{\cal M}}

\newcommand{\boxcmd}{\textbackslash\textbackslash box\{\} }
\newcommand{\boxonly}{\textbackslash\textbackslash box}

\definecolor{agentcolor}{RGB}{0,76,153}   
\definecolor{envcolor}{RGB}{120,120,120}  
\definecolor{highlight}{RGB}{180,0,0}     

\theoremstyle{plain}

\theoremstyle{definition}

\theoremstyle{remark}

\usepackage[textsize=tiny]{todonotes}

\usepackage[square,sort,comma,numbers]{natbib}
\usepackage[preprint]{packages/neurips_2024}

\usepackage[utf8]{inputenc} 
\usepackage[T1]{fontenc}    
\usepackage{hyperref}       
\usepackage{url}            
\usepackage{booktabs}       
\usepackage{amsfonts}       
\usepackage{nicefrac}       
\usepackage{microtype}      
\usepackage{xcolor}         

\title{Scaling In-Context Online Learning Capability of LLMs via Cross-Episode Meta-RL}

\author{
Xiaofeng Lin$^{1}$\thanks{Equal contribution.}\And
Sirou Zhu$^{2}$\footnotemark[1] \And
Yilei Chen$^{1}$ \And
Mingyu Chen$^{1}$ \And
Hejian Sang$^{2}$ \And
Ioannis Paschalidis$^{1}$ \And
Zhipeng Wang$^{2}$\And
Aldo Pacchiano$^{1}$ \And
Xuezhou Zhang$^{1}$ \thanks{Corresponding author: \href{mailto:xuezhouz@bu.edu}{xuezhouz@bu.edu}} 
\\[0.5em]
$^{1}$Boston University, Boston, MA \quad
$^{2}$LinkedIn, Sunnyvale, CA
}

\begin{document}

\maketitle

\begin{abstract}
Large language models (LLMs) achieve strong performance when all task-relevant information is available upfront, as in static prediction and instruction-following problems. However, many real-world decision-making tasks are inherently online: crucial information must be acquired through interaction, feedback is delayed, and effective behavior requires balancing information collection and exploitation over time. While in-context learning enables adaptation without weight updates, existing LLMs often struggle to reliably leverage in-context interaction experience in such settings. In this work, we show that this limitation can be addressed through training. We introduce \textsc{Orbit}, a multi-task, multi-episode meta–reinforcement learning framework that trains LLMs to learn from interaction \emph{in context}. After meta-training, a relatively small open-source model (Qwen3-14B) demonstrates substantially improved in-context online learning on entirely unseen environments, matching the performance of GPT-5.2 and outperforms standard RL fine-tuning by a large margin. Scaling experiments further reveal consistent gains with model size, suggesting significant headroom for learn-at-inference-time decision-making agents. Code reproducing the results in the paper can be found at: \href{https://github.com/XiaofengLin7/ORBIT}{https://github.com/XiaofengLin7/ORBIT}.
\end{abstract}

\section{Introduction}
\begin{figure}[ht]
  \begin{center}
    \centerline{\includegraphics[width=0.7\columnwidth]{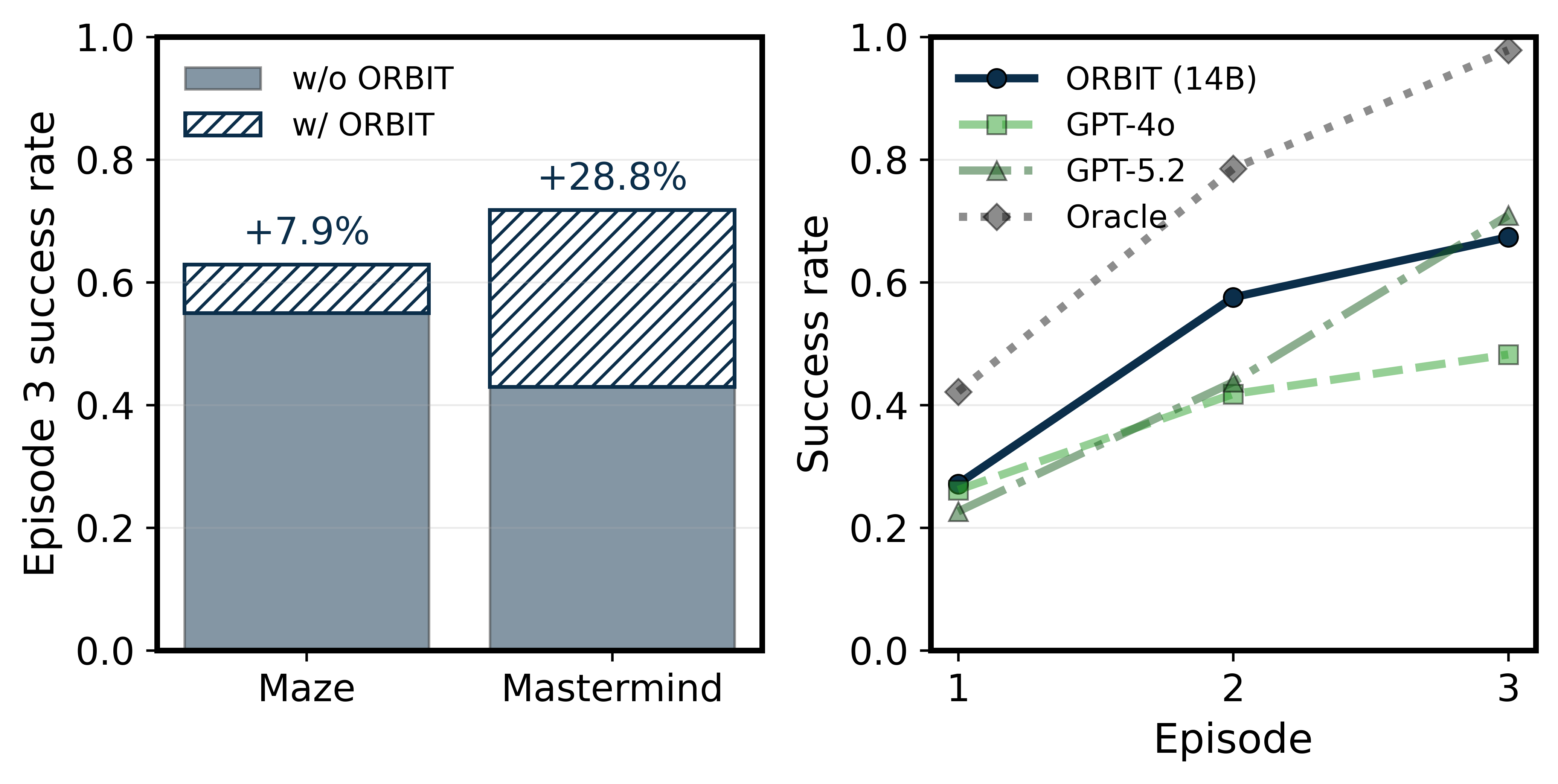}}
    \caption{Left: \textsc{Orbit} gains over the base model on episode‑3 success rate for Maze and Mastermind (w/ \textsc{Orbit} vs. w/o \textsc{Orbit}). Right: Average success rate across Maze and Mastermind over episodes for \textsc{Orbit}, GPT‑4o, GPT‑5.2 (high reasoning effort), and the oracle algorithms for the test environment (details in Appendix \ref{app:oracle}) Both tasks are unseen during training.
    }
    \label{fig:intro}
  \end{center}
\end{figure}
Humans are remarkable learners. While we are far from omniscient, we can often acquire new skills rapidly: a few rounds of trial and error, combined with memory and reasoning, frequently suffice to discover effective strategies in a new game, interface, or problem domain. The capability to use interaction experience to improve behavior over time constitutes a fundamental aspect of intelligence.

In contrast, online learning remains a fundamental challenge for large language models (LLMs). Despite their impressive in-distribution performance, today’s LLMs typically require substantial post-training to reliably adapt to new domains and skills. For example, strong coding performance usually demands explicit training on code, and similarly, domain expertise often follows targeted fine-tuning on domain data. More broadly, once a model is deployed, it rarely improves on a task through continued interaction in the way humans do. This “static-after-shipping” property is a major obstacle for building general-purpose agents that must operate in new environments, recover from mistakes, and refine strategies as they accumulate experience.

While true continual learning through weight updates remains an open problem, a promising alternative is \emph{in-context learning} (ICL), where a model adapts to new information using only its context window. First highlighted in GPT-3 \citep{brown2020gpt3}, ICL shows that pretrained LLMs can condition on demonstrations and feedback without parameter updates. However, the settings that matter for autonomy are not static input--output prediction: they require learning a decision strategy from sequential experience. In partially observed, interactive tasks, an agent must explore to acquire information, perform credit assignment, and then exploit what it has learned, all while the \textit{training signal} arrives as delayed feedback from the environment. Empirically, even frontier models struggle to reliably exploit in-context experience to perform effective online decision making in such settings (Fig.~\ref{fig:intro}), suggesting that strong language modeling alone does not yield robust \emph{in-context online learning}.

To make this concrete, we focus on a particular online learning regime that arises naturally in practice. An agent is deployed to solve an unfamiliar task, such as navigating a new interface, operating in an unknown environment, or interacting with a system whose rules are not fully specified upfront. On its first attempt, the agent may act sensibly yet fail due to hidden constraints or incomplete understanding. Importantly, the task itself does not change: the agent is typically given multiple chances to try again, each starting from a fresh initial state but governed by the same underlying dynamics. A competent learner should therefore treat early attempts as opportunities to gather information and use what it has observed to improve behavior in later ones. We refer to this regime as \emph{multi-episode in-context online learning}, where adaptation must occur across repeated trials of the same task using only the interaction history stored in the context window, without updating model parameters.

This paper asks a simple question:
\begin{quote}
\emph{Can we endow LLMs with general-purpose in-context online learning capability?}
\end{quote}

We answer this in the affirmative by presenting \textsc{Orbit}, a simple yet effective \emph{multi-task, multi-episode meta--reinforcement learning} framework that enables LLMs to perform in-context online learning in unseen tasks. The key idea is to meta-train a pretrained LLM across a diverse distribution of decision-making environments and across multiple episodes per task to maximize long-term reward, for which active information collection becomes a must. In this training setup, the model is not rewarded merely for solving a single instance; it is rewarded for \emph{learning to learn within context}. Early interaction must be used to gather task-relevant information and reduce uncertainty, while later interaction must exploit this information to achieve higher returns. Importantly, we intentionally keep the framework as simple as possible, avoiding additional components such as external memory \citep{packer2023memgpt} or extensive prompt engineering for summarization and reflection \citep{shinn2024reflexion} that have been explored in prior work, in order to isolate and highlight the sole effect of \textbf{multi-episode meta-RL}.

Our results provide strong evidence that such capability can emerge. After meta-training on a suite of partially observable decision-making tasks, a relatively small open-source model (Qwen3-14B) exhibits substantially improved in-context online learning ability on \emph{completely unseen} environments, matching GPT-5.2 and outperforms traditional RL fine-tuning baselines (Fig. \ref{fig:intro} and Fig. \ref{fig:learn_in_ctx}). We further conduct scaling experiments and observe consistent gains as model size increases, suggesting meaningful headroom for future progress. Overall, our findings highlight meta reinforcement learning as a promising pathway toward general-purpose, online decision-making agents that can learn effectively at inference time.

\begin{figure*}[t]
  \begin{center}
    \centerline{\includegraphics[width=0.9\columnwidth]{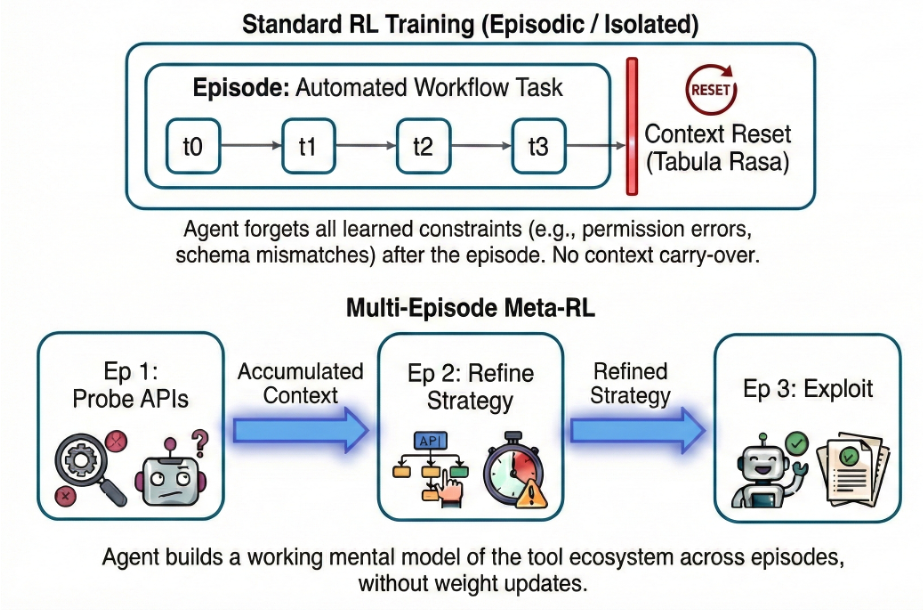}}
    \caption{Example of Multi-Episode Meta-RL for Enterprise Tool-Use. (Top) Standard RL typically treats episodes as isolated events with state resets (tabula rasa), preventing the transfer of learned environmental constraints (e.g., API schemas, rate limits) between trials. (Bottom) A Meta-RL framework enables in-context adaptation across episodes. The agent accumulates a persistent interaction history, allowing it to transition from probing unknown tools (Ep 1) to refining its strategy based on errors (Ep 2), and finally exploiting the learned mental model (Ep 3) for reliable execution, all without requiring weight updates.}
    \label{fig:learn_in_ctx}
  \end{center}
\end{figure*}

\subsection{Related Work}

\paragraph{The In-context Learning Phenomenon}
In-context learning (ICL) was popularized by the GPT-3 technical report \citep{brown2020gpt3}, which demonstrated that large language models can follow task instructions from a small number of prompt demonstrations without parameter updates. Most subsequent studies and practical deployments of ICL focus on few-shot \emph{instruction following} under static input--output formats, often further reinforced by instruction tuning and preference-based post-training methods \citep{christiano2017preferences,ouyang2022instructgpt,rafailov2023dpo}. Mechanistic analyses suggest that ICL may correspond to implicit optimization implemented by the model’s forward pass \citep{vonoswald2023icgd,ahn2023pcgd}. However, these settings typically do not require agents to actively explore or adapt through ongoing interaction, which is the central challenge in continual learning.

\paragraph{In-context Reinforcement Learning}
Parallel to developments in language modeling, the reinforcement learning community has long pursued general-purpose RL algorithms. With the advent of transformers \citep{vaswani2017attention}, numerous works have attempted to train transformers as sequential decision-making agents, most notably Decision Transformer \citep{chen2021decision}. A growing body of work studies in-context reinforcement learning (ICRL) on classical RL benchmarks such as Atari, aiming to train sequence models that adapt online from interaction histories \citep{laskin2023ad,lee2023supervised,lin2024transformers,raparthy2023generalization}. However, many of these approaches train models from scratch or from narrow trajectory distributions, which limits their ability to generalize to unseen tasks. In contrast, starting from a pretrained language model offers a strong semantic and reasoning prior, motivating our focus on general-purpose in-context RL built on top of pretrained LLMs.

\paragraph{In-context Online and Reinforcement Learning in Language Models}
Recent studies have directly evaluated frontier LLMs as in-context online learners on canonical problems such as multi-armed bandits, often finding brittle or inconsistent exploration behavior in the absence of targeted training \citep{krishnamurthy2024explore,nie2024evolve,felicioni2024uncertainty,park2024llm,zhang2025comparing,rahn2024controlling,monea2024llms,sun2025large}. Motivated by these limitations, several recent works aim to improve the in-context RL capabilities of LLMs via post-training on self-generated interaction data. Iterative RMFT \citet{park2025rmft} uses regret-based trajectory selection followed by iterative supervised fine-tuning, but is primarily evaluated in canonical multi-armed bandit settings. PAPRIKA \citep{tajwar2025paprika} constructs preference signals from sampled trajectories for multi-turn post-training and similarly evaluates generalization to new tasks, but reports relatively weak performance on several environments (e.g., $2\%$ success rates on Mastermind). Closest to our approach is the concurrent meta-RL framework of \citet{yan2026pacevolve}, whose objective can be viewed as a discounted variant of ours; however, their evaluation focuses on environments identical to those seen during training, making it difficult to isolate the effect of ICL vs memorizing the environment. To the best of our knowledge, our work is the first to demonstrate substantial and transferable improvements in in-context reinforcement learning capability across diverse, unseen tasks.

\section{In-context Reinforcement Learning}
We model each task as a finite-horizon episodic Markov decision process (MDP)
\[
M = (\mathcal{S}, \mathcal{A}, P, r, \rho, H),
\]
where $\mathcal{S}$ and $\mathcal{A}$ are the state and action spaces, $P(\cdot \mid s,a)$ is the transition kernel, $r(s,a)$ is the reward function, $\rho$ is the initial-state distribution, and $H$ is the (per-episode) horizon. An episode begins with $s_{e,0} \sim \rho$ and proceeds for at most $H$ steps, terminating early if a terminal state is reached.

\paragraph{In-context policies as online RL algorithms.}
We consider an LLM agent with parameters $\theta$ that selects actions using only the interaction transcript stored in its context window. In episode $e \in \{1,\ldots,T\}$ and within-episode timestep $t \in \{0,\ldots,H-1\}$, the agent observes $s_{e,t}$ and samples an action
\[
a_{e,t} \sim \pi_\theta(\cdot \mid h_{e,t}),
\]
where $h_{e,t}$ denotes the full history available in context up to that point. Concretely, we can write
\[
h_{e,t}
=
\bigl(\tau^{(1)}, \ldots, \tau^{(e-1)}, s_{e,0}, a_{e,0}, r_{e,0}, \ldots, s_{e,t}\bigr),
\]
i.e., the concatenation of all prior episodes and the partial trajectory of the current episode. The environment transitions as $s_{e,t+1} \sim P(\cdot \mid s_{e,t}, a_{e,t})$ and emits reward $r_{e,t} = r(s_{e,t}, a_{e,t})$. In this view, an online RL algorithm is simply a (possibly randomized) mapping from histories to action distributions; in-context reinforcement learning aims to realize such an algorithm through the LLM’s forward pass, without any parameter updates at test time.

\paragraph{Multi-episode interaction protocol.}
To capture repeated trials of the \emph{same} underlying task, we evaluate the agent on $T$ episodes of interaction with a fixed MDP $M$. Each episode resets the environment (fresh $s_{e,0} \sim \rho$) while keeping $(P,r)$ unchanged, and the agent retains the entire cross-episode transcript in its context window. We summarize the protocol in Algorithm~\ref{alg:interaction_protocol}. Let $\tau^{(e)}$ denote the trajectory collected in episode $e$:
\[
\tau^{(e)} = (s_{e,0}, a_{e,0}, r_{e,0}, \ldots, s_{e,\ell_e}),
\]
where $\ell_e \le H$ allows for early termination. The full interaction trace across $T$ episodes is the concatenation
$\tau = (\tau^{(1)},\ldots,\tau^{(T)})$.

\begin{algorithm}[tb]
\caption{Multi-Episode Interaction Protocol}
\label{alg:interaction_protocol}
\begin{algorithmic}
\REQUIRE Task MDP $M=(\mathcal{S},\mathcal{A},P,r,\rho,H)$, number of episodes $T$, policy $\pi_\theta$
\ENSURE Cross-episode interaction trace $\tau$
\STATE Initialize empty history $h \leftarrow ()$
\FOR{$e = 1$ ... $T$}
    \STATE $s \leftarrow \texttt{env.reset()}$ \hfill \textit{(sample $s \sim \rho$)}
    \FOR{$t = 0$ ... $H-1$}
        \STATE Sample action $a \sim \pi_\theta(\cdot \mid h, s)$
        \STATE Execute $a$ and observe next state $s'$ and reward $r = r(s,a)$
        \STATE Update history $h \leftarrow (h, s, a, r, s')$
        \STATE $s \leftarrow s'$
        \IF{$s$ is terminal}
            \STATE \textbf{break}
        \ENDIF
    \ENDFOR
\ENDFOR
\STATE \textbf{return} $\tau \equiv h$
\end{algorithmic}
\end{algorithm}

\paragraph{Evaluation via in-context regret across episodes.}
Let the return of episode $e$ be
\[
G^{(e)} \;=\; \sum_{t=0}^{\ell_e-1} r_{e,t},
\qquad \text{with } \ell_e \le H,
\]
and define the total return over $T$ episodes as $\sum_{e=1}^T G^{(e)}$. Let
\[
J^\star(M) \;=\; \max_{\pi} \; \mathbb{E}\!\left[\sum_{t=0}^{H-1} r(s_t,a_t)\right]
\]
denote the optimal expected per-episode return on $M$ (under the same initial-state distribution $\rho$ and dynamics $P$). We define the \emph{in-context regret} after $T$ episodes as
\[
\mathrm{Reg}_T(M)
\;=\;
T \cdot J^\star(M)
\;-\;
\mathbb{E}\!\left[\sum_{e=1}^{T} G^{(e)}\right],
\]
where the expectation is over the environment randomness (initial states and transitions) and the agent’s action sampling.

A strong in-context online learner is one whose regret grows slowly with $T$, indicating that it quickly extracts task-relevant information from early interaction and exploits it to achieve near-optimal performance in later trials.
Minimizing $\mathrm{Reg}_T(M)$ induces an exploration--exploitation trade-off \emph{across episodes}: actions in early episodes may sacrifice immediate reward to reduce uncertainty about the task, while the resulting information—retained in the context window—can be exploited to improve returns in subsequent episodes, all without any parameter updates.

\section{\textsc{Orbit}: Online Reinforcement-Based In-Context Training}
To train a general-purpose in-context learning agent capable of solving diverse tasks on the fly, we train the agent on a collection of tasks (e.g., Minesweeper and Blackjack) and evaluate it on a disjoint set of unseen tasks (e.g., Mastermind and Maze). These tasks differ substantially in their underlying dynamics, observation structures, and optimal solution strategies, such that naïvely transferring a fixed policy or heuristic across tasks is ineffective; success therefore requires the agent to infer task-specific structure and adapt its behavior through in-context interaction, effectively learning how to learn within each new task.

During training, we sample a task \( M \) from the training task set \( \mathcal{M} \). The agent then interacts with \( M \) for \( HT \) steps, generating a trajectory \( \tau\) according to the multi-episode interaction protocol (Alg. \ref{alg:interaction_protocol}) and accumulating reward \( R(M; \tau) \).
This training procedure can be viewed as a meta-level reinforcement learning problem, where the objective is to learn a non-Markovian meta-policy
\[
\Pi : h_t \mapsto a_t,
\]
implemented by the LLM acting purely in context. The meta-learning objective is to maximize the expected normalized cumulative reward—equivalently, to minimize regret—over the task distribution,
\[
\max_{\Pi} \; \mathbb{E}_{M\sim\calM, \Pi}\!\left[ R(M; \tau)\right].
\]
\subsection{Reward Design}
To prevent reward hacking and ensure alignment with task completion, we define rewards at the trajectory level based on the number of successful task completions within a trajectory. For each task \( M \), let \( \mathcal{G}_M \subseteq \mathcal{S} \) denote the set of goal (terminal success) states, and define a success indicator at time \( t \) as
\[
\mathbb{I}_t^{(M)} = \mathbf{1}\{ s_t \in \mathcal{G}_M \}.
\]

We ignore any task-specific process rewards provided by the environment and instead use a unified binary completion reward. This choice avoids scale-induced imbalances across tasks: if different tasks exhibit heterogeneous reward magnitudes, tasks with larger intrinsic reward scales would dominate the optimization signal and contribute disproportionately large gradients. Using a standardized \(0\!-\!1\) completion reward ensures balanced gradient contributions across tasks and focuses learning on successful task completion rather than exploiting task-specific reward shaping.

The trajectory-level reward is then defined as the success count
\[
R(M; \tau) = \sum_{t=0}^{T-1} \mathbb{I}_t^{(M)},
\]
which measures how many times the agent successfully completes the task within the interaction budget \( T \). The corresponding meta-learning objective is to maximize the expected success count over the task distribution,
\[
\max_{\Pi} \; \mathbb{E}_{M \sim \mathcal{M},\, \tau \sim \Pi}\!\left[ R(M; \tau) \right].
\]

This objective encourages agents to reliably complete tasks rather than exploit dense or shaped reward signals, and naturally supports episodic environments in which multiple task completions may occur within a single trajectory.

\subsection{Policy Optimization}

We optimize the meta-policy \( \Pi \) using Group Relative Policy Optimization (GRPO) \citep{shao2024deepseekmath}, a policy-gradient method that operates directly on trajectory-level returns and avoids explicit value-function estimation. This choice is closely aligned with our reward design, which is intrinsically sparse and outcome-driven: reward is assigned only based on task completion within a trajectory, with no intermediate or process-level supervision.

\begin{table}[t]
\caption{Summary of training and test environments. All environments are partially observable.
$H$ denotes the maximum number of steps per episode, and $T$ denotes the number of episodes.
All training tasks use 512 instances, and all test tasks use 256 instances.}
\vspace{0.3cm}
\centering

\small
\setlength{\tabcolsep}{4pt}
\renewcommand{\arraystretch}{1.05}
\begin{tabular}{lclcc}
\toprule
\textbf{Environment} & \textbf{Split} & \textbf{Task Description} & $\boldsymbol{H}$ & $\boldsymbol{T}$ \\
\midrule
RPS         & Train & Infer opponent action distribution and adapt actions to maximize win rate & 5  & 3 \\
Minesweeper & Train & Infer hidden mines and reveal all safe cells under partial observability & 8  & 3 \\
Hangman     & Train & Sequentially guess letters to uncover a hidden word with limited attempts & 10 & 3 \\
Wordle      & Train & Infer a hidden word using structured feedback from each guess & 10 & 3 \\
Blackjack   & Train & Infer deck composition to maximize expected return in a stochastic card game & 4  & 3 \\
Maze        & Test  & Explore an unknown map to reach a goal using local observations only & 9  & 3 \\
Mastermind  & Test  & Infer a hidden ordered sequence using structured feedback & 3  & 3 \\
\bottomrule
\end{tabular}

\label{tab:env_summary}
\end{table}

For each training task \( M \sim \mathcal{M} \), we sample a group of \( K \) trajectories \( \{\tau^{(k)}\}_{k=1}^{K} \) by rolling out the current meta-policy \( \Pi \) in context. Each trajectory \( \tau^{(k)} \) is assigned a trajectory-level reward \( R(M; \tau^{(k)}) \), defined as the total number of successful task completions within the trajectory. GRPO computes a relative advantage for each trajectory by normalizing rewards within the group,
\[
\hat{A}^{(k)} = R(M; \tau^{(k)}) - \frac{1}{K} \sum_{j=1}^{K} R(M; \tau^{(j)}),
\]
which serves as a baseline for variance reduction.

Let \( \Pi_{\text{old}} \) denote the meta-policy before the update. We define the importance sampling ratio
\[
r_t^{(k)} = \frac{\Pi(a_t^{(k)} \mid h_t^{(k)})}{\Pi_{\text{old}}(a_t^{(k)} \mid h_t^{(k)})},
\]
and apply PPO-style clipped updates with asymmetric clipping bounds \( (\epsilon_{\text{low}}, \epsilon_{\text{high}}) \) to stabilize training. The resulting GRPO objective is
\[
\begin{aligned}
&\mathcal{L}_{\mathrm{GRPO}}(\Pi)
=
\mathbb{E}_{M \sim \mathcal{M}}
\Bigg[
\frac{1}{K} \sum_{k=1}^{K} \sum_{t=0}^{T-1}
\\
&
\min\!\left(
r_t^{(k)} \hat{A}^{(k)},
\;\mathrm{clip}\!\left(r_t^{(k)}, 1-\epsilon_{\text{low}}, 1+\epsilon_{\text{high}}\right)\hat{A}^{(k)}
\right)
\Bigg].
\end{aligned}
\]

A key property of GRPO is that it operates fundamentally at the level of complete trajectories and does not maintain or learn an explicit value function. Consequently, it cannot reliably exploit fine-grained or process-level reward signals that depend on intermediate states or partial progress within a trajectory. Since group sampling, reward normalization, and advantage computation are all performed over full trajectories, the learning signal depends only on relative differences in aggregated trajectory outcomes. In this sense, GRPO is inherently outcome-driven: it compares trajectories based on overall success rather than intermediate feedback.

While the inability to leverage dense process rewards may be a limitation in settings where such signals are informative, it is well matched to our objective. Our reward design is intentionally sparse and aligned with task completion, both to avoid reward hacking and to ensure balanced optimization across heterogeneous tasks. Moreover, recent empirical evidence suggests that suppressing step-wise credit assignment can be beneficial for reasoning-intensive settings. For example, the Kimi~K1.5 technical report \citep{team2025kimi} observes that removing value-function-based credit assignment encourages exploration of diverse reasoning paths, allowing models to recover from intermediate errors and learn effective trial-and-error strategies from final outcomes alone. Although our setting differs, this observation provides supporting intuition that outcome-level supervision can promote richer in-context adaptation rather than prematurely constraining behavior based on local signals.

\section{Experiments}
\subsection{Environments}

Table~\ref{tab:env_summary} summarizes all training and test environments along with concise task descriptions. All environments are partially observable in the sense that the environment parameters, such as the map of the maze, is initially unknown to the agent. Successful completion of the task requires the agent to balance exploration and exploitation. Detailed prompt designs and task specifications are provided in Appendix~\ref{app:prompts} and Appendix~\ref{app:tasks}.

\subsection{Experimental Setup}
For most experiments in this paper, we use Qwen3-8B as the base model \citep{yang2025qwen3}. In the scaling study, we additionally train agents on top of Qwen3-\{4B, 8B, 14B\} to examine how in-context learning performance scales with model size. All base models have a maximum context length of 32,768 tokens. All experiments are implemented using the RLLM training framework \citep{rllm2025}.

The relatively limited context length of 32k tokens is a primary factor constraining the interaction horizon in our experiments. To ensure that complete interaction histories can be preserved within the context window, we restrict each task instance to three episodes per game. Extending this framework to longer horizons is a natural direction for future work, particularly with larger-context or memory-augmented models.

\textbf{Training setup.} During training, we use temperature \(1.0\) and top-\(p = 1.0\). Trajectories that exceed this limit are truncated and assigned zero reward. For each training instance, we generate \(4\) trajectories and use a batch size of \(64\) across all model sizes. Models are trained for \(100\) optimization steps with a learning rate of \(1 \times 10^{-6}\), using PPO-style clipped policy updates with asymmetric clipping bounds \((0.2, 0.28)\). Both entropy regularization and KL regularization are disabled. We enable the model’s thinking mode during training. Due to limited computational resources, all comparison experiments are conducted using Qwen3-8B; Only scaling law experiments include the 4B and 14B. 

\textbf{Training setup.} During training, we use temperature \(1.0\) and top-\(p = 1.0\). Each trajectory is generated subject to the 32k-token context limit; trajectories that exceed this limit are truncated and assigned zero reward. For each training instance, we sample a group of \(K=4\) trajectories for GRPO optimization. We use a batch size of \(64\) trajectories across all model sizes.

Models are trained for \(100\) optimization steps with a learning rate of \(1 \times 10^{-6}\), using PPO-style clipped policy updates with asymmetric clipping bounds \((0.2, 0.28)\). Both entropy regularization and KL regularization are disabled. We enable the model’s thinking mode during training to match the inference-time configuration.

\textbf{Evaluation setup.} For Sections~\ref{sec:orbit_learn} and~\ref{sec:explore}, we report results from the checkpoint obtained after 100 optimization steps. For Section~\ref{sec:scaling}, we instead report the best-performing checkpoint within the first 100 optimization steps, in order to reflect the full performance potential of \textsc{Orbit}. Model checkpoints are evaluated using the inference settings recommended for Qwen3 models, with temperature \(0.6\) and top-\(p = 0.95\). Thinking mode is enabled during evaluation to match the training configuration.

All evaluations are conducted on two test environments, \emph{Maze} and \emph{Mastermind}, which are entirely unseen during training and each comprise 256 distinct instances. This evaluation protocol is designed to assess generalization beyond the training tasks.
We intentionally do not evaluate on training tasks or on intra-environment variants (e.g., Minesweeper with a larger board or Wordle with longer words), as success rates in these settings typically saturate quickly and make it difficult to disentangle genuine in-context adaptation from environment memorization. Focusing on entirely unseen task classes therefore provides a cleaner evaluation of generalizable in-context reinforcement learning behavior.

\subsection{\textsc{Orbit} Learns In Context}\label{sec:orbit_learn}
\begin{figure}[t]
  \begin{center}
    \centerline{\includegraphics[width=0.7\columnwidth]{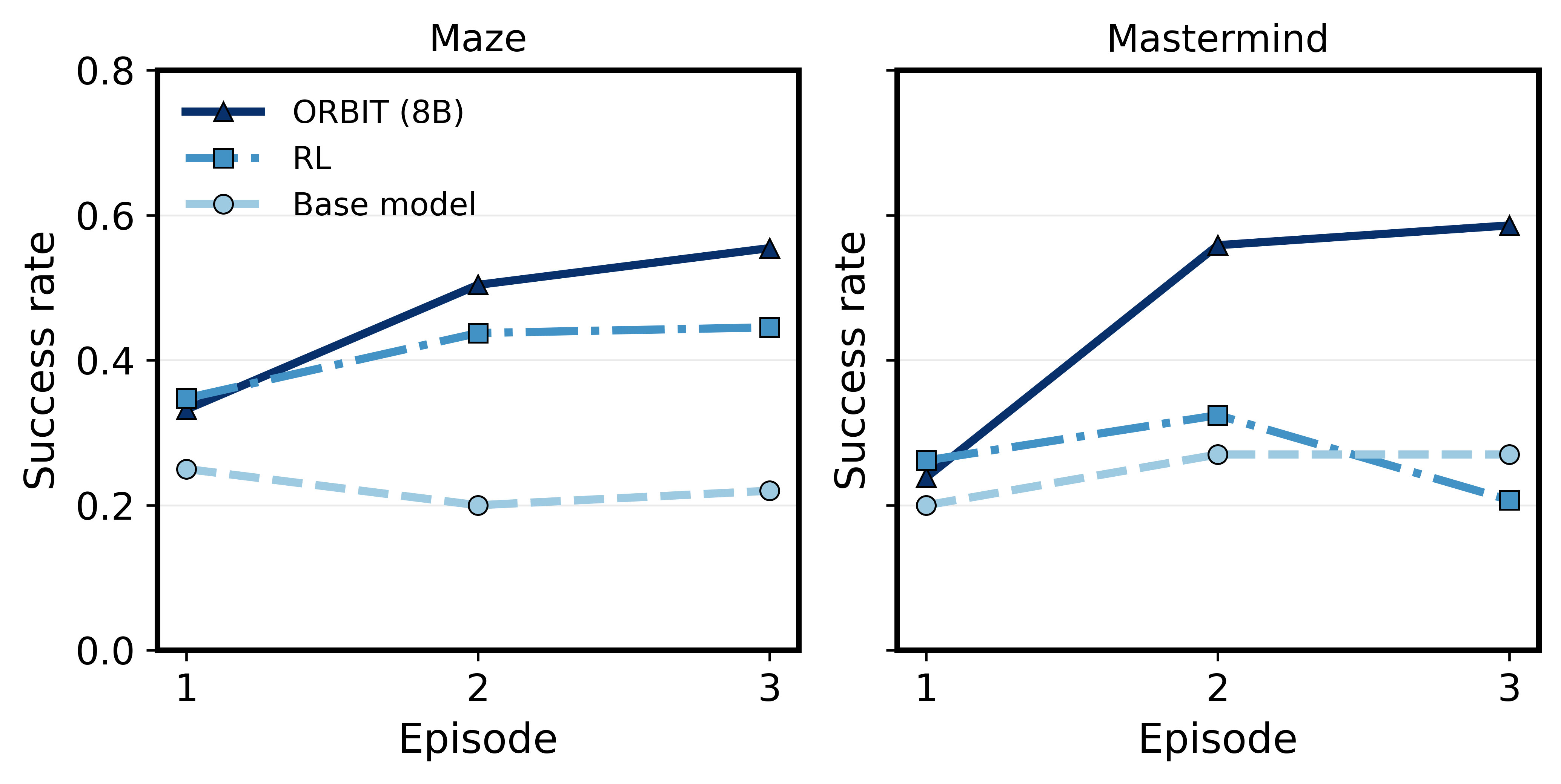}}
    \caption{\textsc{Orbit} induces genuine in-context learning beyond single-episode and multitask baselines. Success rate versus episode index on unseen Maze (left) and Mastermind (right) tasks. We compare \textsc{Orbit} (8B) against the base model and RL baseline. 
    }
    \label{fig:learn_in_ctx}
  \end{center}
\end{figure}

Figure~\ref{fig:learn_in_ctx} shows that \textsc{Orbit} induces genuine in-context learning on the unseen \emph{Maze} and \emph{Mastermind} tasks. Compared to the base model, \textsc{Orbit} (8B) achieves substantially higher success rates and, critically, continues to improve across successive episodes. This consistent upward trend indicates that the model effectively leverages interaction history stored in the context window to refine its behavior online.

Crucially, \textsc{Orbit} also consistently outperforms the standard RL post-training baseline, which is trained on the same task distribution but optimized using single-episode completion rewards. While this RL baseline yields modest improvements over the base model in early episodes, its performance quickly saturates or even degrades as episodes progress, suggesting a limited capacity for in-context adaptation. In contrast, \textsc{Orbit} exhibits monotonic improvement across episodes on both tasks, highlighting its ability to adapt behavior based on feedback from previous interactions.

Table~\ref{tab:ep3_improvement} further reinforces this distinction by focusing on performance in the third episode, after the agent has accumulated interaction history within the same task instance. \textsc{Orbit} consistently outperforms both the base model and the RL baseline, whereas the RL baseline shows limited or unstable gains once early episodes fail. This pattern suggests that \textsc{Orbit}’s improvements are not merely the result of stronger static task-solving policies learned during training, but instead arise from its ability to adapt its behavior within a task using information from prior episodes.

Taken together, these results demonstrate that \textsc{Orbit} learns an in-context policy that supports online adaptation at inference time, enabling exploration and refinement across episodes without parameter updates. This behavior distinguishes \textsc{Orbit} from conventional RL post-training approaches that operate on isolated, single-episode rollouts, and underscores the effectiveness of our framework for inducing in-context reinforcement learning in large language models.

\subsection{\textsc{Orbit} Learns to Explore}\label{sec:explore}
\begin{table}[t]
\caption{\textbf{Episode~3 success rate improvement over the base model.}
We report success rates in Episode~3 on Maze and Mastermind, together with absolute improvements over the base model.}
\vspace{0.5cm}
\centering
\small
\setlength{\tabcolsep}{6pt}
\renewcommand{\arraystretch}{1.15}
\begin{tabular}{lccc}
\toprule
\textbf{Task} & \textbf{Method} & \textbf{Ep~3 Success} & \textbf{$\Delta$ vs. Base} \\
\midrule
\multirow{2}{*}{Maze} 
& RL        & $0.45$ & $+0.23$ \\
& \textsc{Orbit}       & $\mathbf{0.55}$ & $\mathbf{+0.33}$ \\
\midrule
\multirow{2}{*}{Mastermind}
& RL        & $0.21$ & $-0.06$ \\
& \textsc{Orbit}       & $\mathbf{0.59}$ & $\mathbf{+0.32}$ \\
\bottomrule
\end{tabular}
\label{tab:ep3_improvement}
\end{table}

\begin{figure}[ht]
  \begin{center}
    \centerline{\includegraphics[width=0.5\columnwidth]{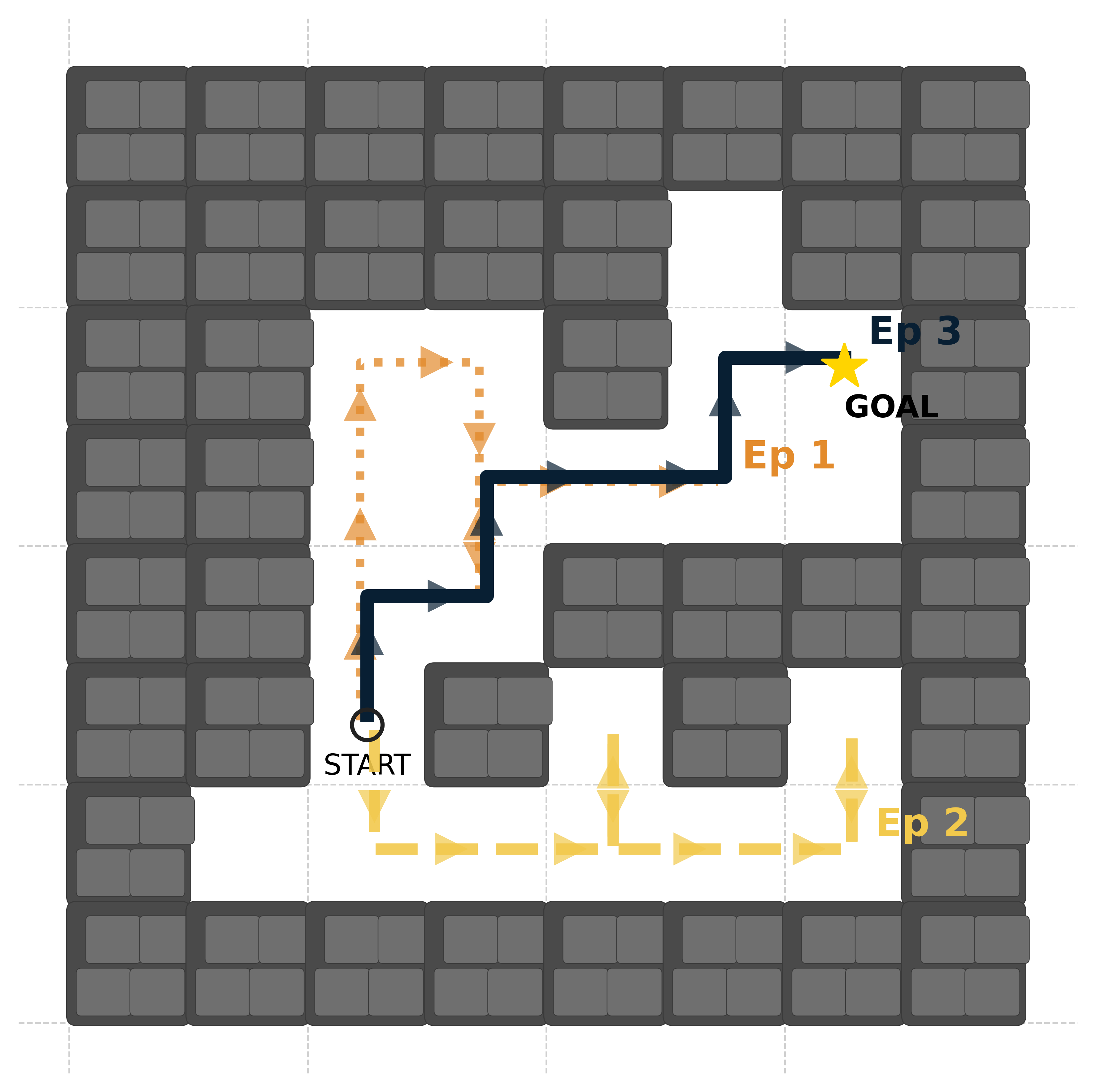}}
    \caption{A trace of \textsc{Orbit} in Maze. The agent observes only its local surroundings and must rely on its history to guide decisions. The first two episodes (Ep 1 and Ep 2) fail to reach the goal, exploring suboptimal paths. Leveraging reflections over past interactions, the agent adapts its strategy and, in the third episode (Ep 3), follows a new route that successfully reaches the goal.
    }
    \label{fig:maze_trace}
  \end{center}
\end{figure}

\begin{table}[tb]
\caption{\textbf{Qualitative in-context “reflect and adapt” moment under \textsc{Orbit}.} The agent fails in the first two episodes. In the third episode, it explicitly summarizes the failures from earlier interactions, reflects on its accumulated history, and selects actions that differ from previous attempts.}
\vspace{0.5cm}
\centering
\small
\setlength{\tabcolsep}{6pt}
\renewcommand{\arraystretch}{1.25}
\begin{tabularx}{\linewidth}{@{}lX@{}}
\toprule
\textbf{Env} &
{\color{envcolor}
\textbf{New episode begins.} You are a maze-solving agent. Your goal is to navigate from the START position to the GOAL position in the fewest turns possible. You are at the START position \((4,1)\). Around you, up leads to path, down leads to path, left leads to wall, and right leads to wall. Output your next move from up/down/left/right within \(\backslash\)boxed\{\}.}
\\
\midrule
\textbf{\textsc{Orbit}} &
{\color{agentcolor}
\texttt{<think>} \ldots
In Episode~1, I went up three times, then right and down, but got stuck. In Episode~2, I went down first and explored the bottom corridor, but that also led to a loop.
\texttt{</think>}
\(\boxed{\text{up}}\)}
\\
\midrule
\textbf{\textsc{Orbit}} &
{\color{agentcolor}
\texttt{<think>} \ldots
Previously, I went up again from here and failed to reach the goal.
This time, I will explore right instead.
\texttt{</think>}
\(\boxed{\text{right}}\)}
\\
\midrule
\textbf{} &
{\color{envcolor}
\textit{[agent moves up, right, right, up, right]}}
\\
\midrule
\textbf{Env} &
{\color{envcolor}
\textbf{Congratulations!} You arrived at the goal.}
\\
\bottomrule
\end{tabularx}

\label{tab:orbit_aha_trace}

\end{table}
\begin{table}[t]
\caption{\textbf{Exploration under failure.} 
\(\Delta\)States (Ep2$|$F1) denotes the number of newly explored states in Episode~2 conditioned on failure in Episode~1, and 
\(\Delta\)States (Ep3$|$F1--2) denotes the number of newly explored states in Episode~3 conditioned on failure in both Episodes~1 and~2. Results are collected using Qwen3-8B.}
\vspace{0.5cm}
\centering

\small
\setlength{\tabcolsep}{6pt}
\renewcommand{\arraystretch}{1.15}
\begin{tabular}{lcc}
\toprule
\textbf{Method} & \(\Delta\)States (Ep2$|$F1) & \(\Delta\)States (Ep3$|$F1--2) \\
\midrule
Base  &    $1.64$    &   $0.94$  \\
RL  & $4.11$ & $1.27$ \\
\textbf{\textsc{Orbit}} & $\mathbf{4.69}$ & $\mathbf{1.48}$ \\
\bottomrule
\end{tabular}
\label{tab:exploration_comparison}
\end{table}

Figure~\ref{fig:maze_trace} and, in particular, the snippet in Table~\ref{tab:orbit_aha_trace} provide qualitative evidence that \textsc{Orbit} acquires an exploration strategy \emph{through in-context adaptation} rather than relying on fixed heuristics or hand-engineered prompting. In the partially observable maze, the agent observes only local surroundings and must use the cross-episode transcript as its memory. After failing in Episodes~1--2, \textsc{Orbit} enters Episode~3 and spontaneously performs behaviors commonly associated with deliberative exploration: it \emph{reflects} on what went wrong, summarizes the relevant parts of the interaction history, and then chooses actions designed to gather new information instead of replaying the same trajectory.

Crucially, these behaviors are not prescribed by an explicit ``reflect'' instruction or an auxiliary module: they emerge from optimizing the multi-episode meta-RL objective alone. When faced with the same local observation, \textsc{Orbit} selects actions that intentionally differ from earlier episodes (Table~\ref{tab:orbit_aha_trace}), indicating genuinely history-dependent decision making. This shift leads the agent to actively explore a previously unvisited route and successfully reach the goal (Fig.~\ref{fig:maze_trace}).

Beyond the qualitative trace, Table~\ref{tab:exploration_comparison} provides quantitative evidence that \textsc{Orbit} changes its exploration behavior \emph{conditioned on failure}. We measure how many \emph{new} states are visited in later episodes given that the agent failed earlier: \(\Delta\)States (Ep2$|$F1) counts newly explored states in Episode~2 conditioned on failure in Episode~1, while \(\Delta\)States (Ep3$|$F1--2) counts newly explored states in Episode~3 conditioned on failure in both Episodes~1 and~2. Under this conditioning, an effective online learner should avoid repeating unproductive behaviors and instead actively expand coverage of the state space to reduce uncertainty. \textsc{Orbit} consistently explores more new states than the single-episode RL baseline, suggesting that it learns to ``try something different'' after failure rather than cycling through the same local choices.

Taken together, the qualitative evidence (Table~\ref{tab:orbit_aha_trace}, Fig.~\ref{fig:maze_trace}) and the quantitative analysis (Table~\ref{tab:exploration_comparison}) indicate that \textsc{Orbit} learns an adaptive exploration policy in-context: it can perform reflection-like summarization and active exploration strategies on its own, using only the interaction history stored in the context window, and improves across episodes without any parameter updates at inference time.

\subsection{\textsc{Orbit} Scales with Model Size}\label{sec:scaling}

\begin{figure}[t]
  \begin{center}
    \centerline{\includegraphics[width=0.7\columnwidth]{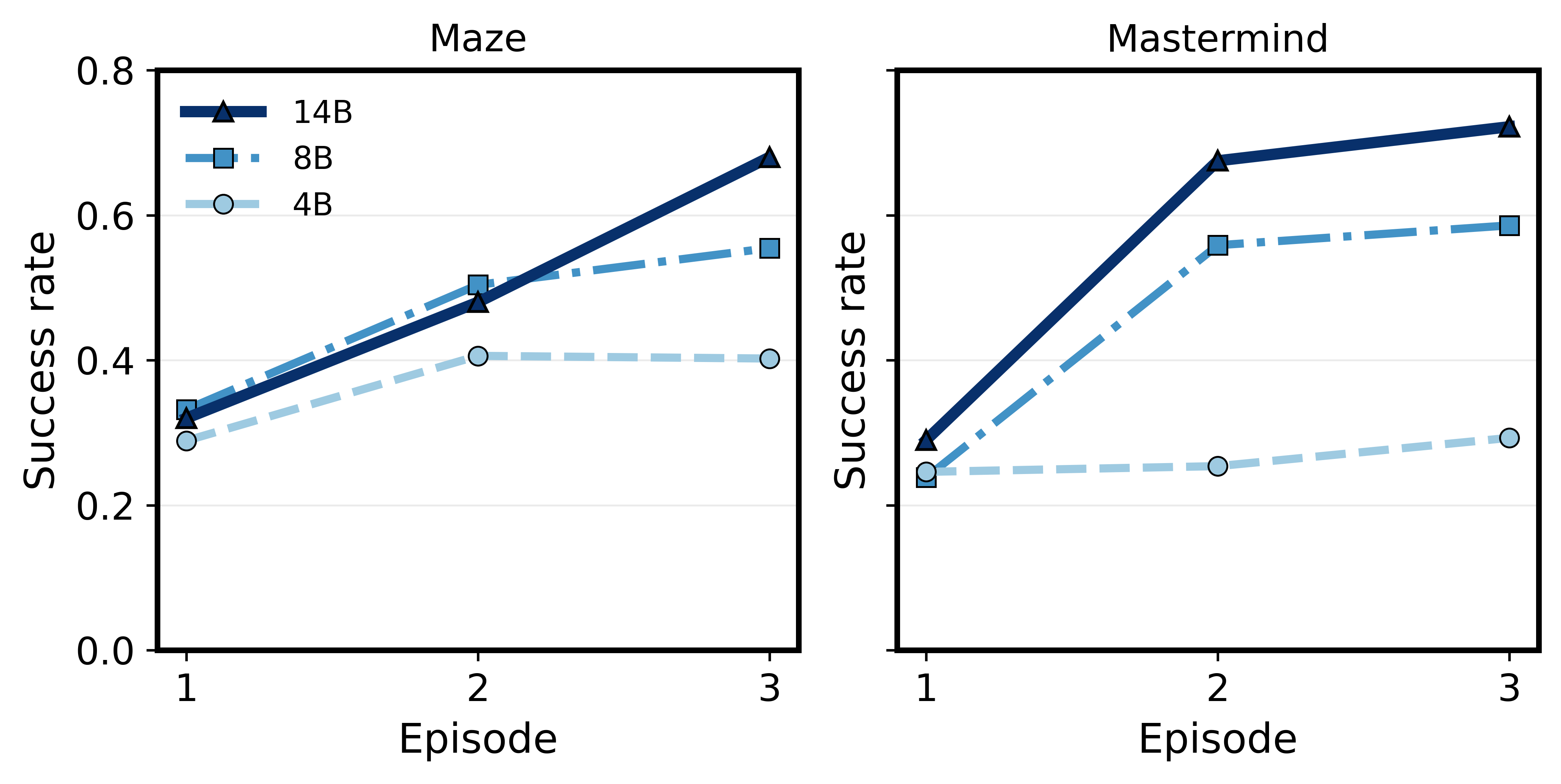}}
    \caption{In-context learning across episodes on unseen test tasks. Success rate versus episode index for Maze (left) and Mastermind (right) when evaluating Qwen3-{4B, 8B, 14B} trained with \textsc{Orbit}. We report the best-performing checkpoint in terms of success rate of episode 3 within the first 100 training steps to highlight \textsc{Orbit}’s potential.
    }
    \label{fig:scaling_law}
  \end{center}
\end{figure}

Figure~\ref{fig:scaling_law} reveals a consistent benefit from scaling the Qwen3 backbone from 4B to 8B to 14B on the unseen Maze and Mastermind benchmarks, with the largest gains appearing in later episodes. In particular, Episode~3 improves the most with model size, followed by Episode~2, while Episode~1 shows only modest improvement (and can even drop for 14B).

This pattern is consistent with the interpretation that larger \textsc{Orbit} models may allocate early interaction more toward information gathering (exploration) rather than immediate task completion (exploitation), and then capitalize on the accumulated cross-episode evidence in later attempts. Overall, scaling model capacity appears to amplify---rather than replace---the multi-episode in-context learning dynamics induced by \textsc{Orbit}.

\section{Discussion and Future Work}
This work primarily demonstrates the feasibility of using meta reinforcement learning as a mechanism for inducing generalizable ICRL capabilities, while leaving substantial room for further improvement. Importantly, we intentionally keep the framework as simple as possible, avoiding additional components such as external memory, retrieval-augmented generation, or extensive prompt engineering for summarization and reflection that have been explored in prior work, in order to isolate and highlight the contribution of \textbf{multi-episode meta learning} alone. Our experiments are limited to relatively short interaction horizons due to the 32k-token context length of the Qwen-3 base model; extending training to longer horizons, either with larger-context models or memory-augmented architectures, would enable richer temporal reasoning and more effective long-term credit assignment. In addition, we train on only five environments, and scaling both the number and diversity of training environments is a natural next step toward understanding empirical scaling laws governing generalization. There is also significant potential to improve the computational efficiency and stability of the training procedure through more advanced optimization and credit-assignment techniques.

\bibliography{ref}

@article{monea2024llms,
  title={LLMs Are In-Context Bandit Reinforcement Learners},
  author={Monea, Giovanni and Bosselut, Antoine and Brantley, Kiant{\'e} and Artzi, Yoav},
  journal={arXiv preprint arXiv:2410.05362},
  year={2024}
}

@article{chen2021decision,
  title={Decision transformer: Reinforcement learning via sequence modeling},
  author={Chen, Lili and Lu, Kevin and Rajeswaran, Aravind and Lee, Kimin and Grover, Aditya and Laskin, Misha and Abbeel, Pieter and Srinivas, Aravind and Mordatch, Igor},
  journal={Advances in neural information processing systems},
  volume={34},
  pages={15084--15097},
  year={2021}
}

@article{sun2025large,
  title={Large language model-enhanced multi-armed bandits},
  author={Sun, Jiahang and Wang, Zhiyong and Yang, Runhan and Xiao, Chenjun and Lui, John and Dai, Zhongxiang},
  journal={arXiv preprint arXiv:2502.01118},
  year={2025}
}

@article{team2025kimi,
  title={Kimi k1. 5: Scaling reinforcement learning with llms},
  author={Team, Kimi and Du, Angang and Gao, Bofei and Xing, Bowei and Jiang, Changjiu and Chen, Cheng and Li, Cheng and Xiao, Chenjun and Du, Chenzhuang and Liao, Chonghua and others},
  journal={arXiv preprint arXiv:2501.12599},
  year={2025}
}

@article{rahn2024controlling,
  title={Controlling large language model agents with entropic activation steering},
  author={Rahn, Nate and D'Oro, Pierluca and Bellemare, Marc G},
  journal={arXiv preprint arXiv:2406.00244},
  year={2024}
}

@article{zhang2025comparing,
  title={Comparing Exploration-Exploitation Strategies of LLMs and Humans: Insights from Standard Multi-armed Bandit Tasks},
  author={Zhang, Ziyuan and Wang, Darcy and Chen, Ningyuan and Mansur, Rodrigo and Sarhangian, Vahid},
  journal={arXiv preprint arXiv:2505.09901},
  year={2025}
}

@article{yang2025qwen3,
  title={Qwen3 technical report},
  author={Yang, An and Li, Anfeng and Yang, Baosong and Zhang, Beichen and Hui, Binyuan and Zheng, Bo and Yu, Bowen and Gao, Chang and Huang, Chengen and Lv, Chenxu and others},
  journal={arXiv preprint arXiv:2505.09388},
  year={2025}
}

@inproceedings{park2024llm,
  title={Do llm agents have regret? a case study in online learning and games},
  author={Park, Chanwoo and Liu, Xiangyu and Ozdaglar, Asuman and Zhang, Kaiqing},
  booktitle={International Conference on Learning Representations},
  year={2025}
}

@article{shao2024deepseekmath,
  title={Deepseekmath: Pushing the limits of mathematical reasoning in open language models},
  author={Shao, Zhihong and Wang, Peiyi and Zhu, Qihao and Xu, Runxin and Song, Junxiao and Zhang, Mingchuan and Li, YK and Wu, Yu and Guo, Daya},
  journal={arXiv preprint arXiv:2402.03300},
  year={2024}
}

@article{vaswani2017attention,
  title={Attention is all you need},
  author={Vaswani, Ashish and Shazeer, Noam and Parmar, Niki and Uszkoreit, Jakob and Jones, Llion and Gomez, Aidan N and Kaiser, {\L}ukasz and Polosukhin, Illia},
  journal={Advances in neural information processing systems},
  volume={30},
  year={2017}
}

@article{lee2023supervised,
  title={Supervised Pretraining Can Learn In-Context Reinforcement Learning},
  author={Lee, Jonathan N and Xie, Annie and Pacchiano, Aldo and Chandak, Yash and Finn, Chelsea and Nachum, Ofir and Brunskill, Emma},
  journal={Neural Information Processing
Systems},
  year={2023}
}

@article{shinn2024reflexion,
  title={Reflexion: Language agents with verbal reinforcement learning},
  author={Shinn, Noah and Cassano, Federico and Gopinath, Ashwin and Narasimhan, Karthik and Yao, Shunyu},
  journal={Advances in Neural Information Processing Systems},
  volume={36},
  year={2024}
}

@article{packer2023memgpt,
  title={MemGPT: Towards LLMs as Operating Systems.},
  author={Packer, Charles and Fang, Vivian and Patil, Shishir\_G and Lin, Kevin and Wooders, Sarah and Gonzalez, Joseph\_E},
  year={2023},
  publisher={ArXiv}
}

@inproceedings{
lin2024transformers,
title={Transformers as Decision Makers: Provable In-Context Reinforcement Learning via Supervised Pretraining},
author={Licong Lin and Yu Bai and Song Mei},
booktitle={The Twelfth International Conference on Learning Representations},
year={2024},
url={https://openreview.net/forum?id=yN4Wv17ss3}
}

@article{raparthy2023generalization,
  title={Generalization to new sequential decision making tasks with in-context learning},
  author={Raparthy, Sharath Chandra and Hambro, Eric and Kirk, Robert and Henaff, Mikael and Raileanu, Roberta},
  journal={arXiv preprint arXiv:2312.03801},
  year={2023}
}

@inproceedings{brown2020gpt3,
  title     = {Language Models are Few-Shot Learners},
  author    = {Brown, Tom B. and Mann, Benjamin and Ryder, Nick and Subbiah, Melanie and Kaplan, Jared and Dhariwal, Prafulla and Neelakantan, Arvind and Shyam, Pranav and Sastry, Girish and Askell, Amanda and Agarwal, Sandhini and Herbert-Voss, Ariel and Krueger, Gretchen and Henighan, Tom and Child, Rewon and Ramesh, Aditya and Ziegler, Daniel M. and Wu, Jeffrey and Winter, Clemens and Hesse, Christopher and Chen, Mark and Sigler, Eric and Litwin, Mateusz and Gray, Scott and Chess, Benjamin and Clark, Jack and Berner, Christopher and McCandlish, Sam and Radford, Alec and Sutskever, Ilya and Amodei, Dario},
  booktitle = {Advances in Neural Information Processing Systems (NeurIPS)},
  year      = {2020},
  url       = {https://arxiv.org/abs/2005.14165}
}

@article{yan2026pacevolve,
  title={PACEvolve: Enabling Long-Horizon Progress-Aware Consistent Evolution},
  author={Yan, Minghao and Peng, Bo and Coleman, Benjamin and Chen, Ziqi and Xie, Zhouhang and He, Zhankui and Sachdeva, Noveen and Ye, Isabella and Wang, Weili and Wang, Chi and others},
  journal={arXiv preprint arXiv:2601.10657},
  year={2026}
}

@article{tajwar2025paprika,
  title   = {Training a Generally Curious Agent},
  author  = {Tajwar, Fahim and Jiang, Yiding and Thankaraj, Abitha and Rahman, Sumaita Sadia and Kolter, J. Zico and Schneider, Jeff and Salakhutdinov, Ruslan},
  journal = {arXiv preprint arXiv:2502.17543},
  year    = {2025},
  url     = {https://arxiv.org/abs/2502.17543}
}

@article{park2025rmft,
  title   = {Post-Training {LLMs} as Better Decision-Making Agents: A Regret-Minimization Approach},
  author  = {Park, Chanwoo and Chen, Ziyang and Ozdaglar, Asuman and Zhang, Kaiqing},
  journal = {arXiv preprint arXiv:2511.04393},
  year    = {2025},
  url     = {https://arxiv.org/abs/2511.04393}
}

@article{krishnamurthy2024explore,
  title   = {Can Large Language Models Explore In-Context?},
  author  = {Krishnamurthy, Akshay and Harris, Keegan and Foster, Dylan J. and Zhang, Cyril and Slivkins, Aleksandrs},
  journal = {arXiv preprint arXiv:2403.15371},
  year    = {2024},
  url     = {https://arxiv.org/abs/2403.15371}
}

@article{nie2024evolve,
  title   = {{EVOLvE}: Evaluating and Optimizing {LLMs} For Exploration},
  author  = {Nie, Allen and Su, Yi and Chang, Bo and Lee, Jonathan N. and Chi, Ed H. and Le, Quoc V. and Chen, Minmin},
  journal = {arXiv preprint arXiv:2410.06238},
  year    = {2024},
  url     = {https://arxiv.org/abs/2410.06238}
}

@article{christiano2017preferences,
  title   = {Deep Reinforcement Learning from Human Preferences},
  author  = {Christiano, Paul F. and Leike, Jan and Brown, Tom B. and Martic, Miljan and Legg, Shane and Amodei, Dario},
  journal = {arXiv preprint arXiv:1706.03741},
  year    = {2017},
  url     = {https://arxiv.org/abs/1706.03741}
}

@article{ouyang2022instructgpt,
  title   = {Training Language Models to Follow Instructions with Human Feedback},
  author  = {Ouyang, Long and Wu, Jeff and Jiang, Xu and Almeida, Diogo and Wainwright, Carroll L. and Mishkin, Pamela and Zhang, Chong and Agarwal, Sandhini and Slama, Katarina and Ray, Alex and Schulman, John and Hilton, Jacob and Kelton, Fraser and Miller, Luke and Simens, Maddie and Askell, Amanda and Welinder, Peter and Christiano, Paul F. and Leike, Jan and Lowe, Ryan},
  journal = {arXiv preprint arXiv:2203.02155},
  year    = {2022},
  url     = {https://arxiv.org/abs/2203.02155}
}

@inproceedings{rafailov2023dpo,
  title     = {Direct Preference Optimization: Your Language Model is Secretly a Reward Model},
  author    = {Rafailov, Rafael and Sharma, Archit and Mitchell, Eric and Ermon, Stefano and Manning, Christopher D. and Finn, Chelsea},
  booktitle = {Advances in Neural Information Processing Systems (NeurIPS)},
  year      = {2023},
  url       = {https://arxiv.org/abs/2305.18290}
}

@article{felicioni2024uncertainty,
  title   = {On the Importance of Uncertainty in Decision-Making with Large Language Models},
  author  = {Felicioni, Nicol{\`o} and Maystre, Lucas and Ghiassian, Seth and Ciosek, Kamil},
  journal = {arXiv preprint arXiv:2404.02649},
  year    = {2024},
  url     = {https://arxiv.org/abs/2404.02649}
}

@inproceedings{laskin2023ad,
  title     = {In-context Reinforcement Learning with Algorithm Distillation},
  author    = {Laskin, Michael and Wang, Luyu and Oh, Junhyuk and Parisotto, Emilio and Spencer, Stephen and Steigerwald, Richie and Strouse, DJ and Hansen, Steven and Filos, Angelos and Brooks, Ethan and Gazeau, Maxime and Sahni, Himanshu and Singh, Satinder and Mnih, Volodymyr},
  booktitle = {International Conference on Learning Representations (ICLR)},
  year      = {2023},
  url       = {https://arxiv.org/abs/2210.14215}
}

@inproceedings{vonoswald2023icgd,
  title     = {Transformers Learn In-Context by Gradient Descent},
  author    = {von Oswald, Johannes and Niklasson, Eyvind and Randazzo, Emanuele and Sacramento, Jo{\~a}o and Mordvintsev, Alexander and Zhmoginov, Andrey and Vladymyrov, Max},
  booktitle = {International Conference on Machine Learning (ICML)},
  year      = {2023}
}

@inproceedings{ahn2023pcgd,
  title     = {Transformers Learn to Implement Preconditioned Gradient Descent for In-Context Learning},
  author    = {Ahn, Kwangjun and Cheng, Xinyi and Daneshmand, Hadi and Sra, Suvrit},
  booktitle = {Advances in Neural Information Processing Systems (NeurIPS)},
  year      = {2023}
}

@misc{rllm2025,
  title={rLLM: A Framework for Post-Training Language Agents},
  author={Sijun Tan and Michael Luo and Colin Cai and Tarun Venkat and Kyle Montgomery and Aaron Hao and Tianhao Wu and Arnav Balyan and Manan Roongta and Chenguang Wang and Li Erran Li and Raluca Ada Popa and Ion Stoica},
  year={2025},
  howpublished={\url{https://pretty-radio-b75.notion.site/rLLM-A-Framework-for-Post-Training-Language-Agents-21b81902c146819db63cd98a54ba5f31}},
  note={Notion Blog},
  year={2025}
}
\bibliographystyle{plainnat}

\newpage
\appendix

\section{Prompts}\label{app:prompts}

\begin{tcolorbox}[colback=gray!10,colframe=black,title={System Prompt}]
You are solving the same task across multiple episodes with a fixed total step budget. Each episode resets the environment but keeps the task identical. Leverage information gathered from earlier episodes to succeed faster. Respond with actions inside \boxcmd each turn.
\end{tcolorbox}

\begin{tcolorbox}
[colback=gray!10,colframe=black,title={Prompt for Minesweeper}]
You are playing the Minesweeper game. The objective of the game is to reveal all cells that do not contain mines. To make a move, you can either reveal a cell or place a flag on a suspected mine location using one of the following commands: \\
- 'reveal': Reveal the contents of a specific cell. \\
- 'flag': Place or remove a flag on a specific cell to mark it as a potential mine.\\
To submit your move, type the command followed by the row and column in \boxcmd. \\
For example:\\
- \boxonly\{reveal 4 4\} to reveal the cell in Row 4, Column 4. \\
- \boxonly\{flag 2 2\} to place or remove a flag on the cell in Row 2, Column 2.\\
The current board layout is shown below. Cells that are unrevealed are represented by a dot ('.'), revealed numbers show the count of adjacent mines, and flagged cells are marked with an 'F'.\\
Use logic and deduction to avoid revealing cells with mines! Be mindful not to choose revealed or flagged cells. \\
Here is the current board layout:
\begin{verbatim}
    0  1  2  3  4
 0  .  .  .  .  .
 1  .  .  .  .  .
 2  .  .  .  .  .
 3  .  .  .  .  .
 4  .  .  .  .  .
\end{verbatim}
Enter your guess.
\end{tcolorbox}

\begin{tcolorbox}[colback=gray!10,colframe=black,title={Prompt for Hangman}]
You are playing Hangman.
The objective of the game is to guess the 3-letter word by providing one letter guesses or the entire word.\\
The cells that need to be populated with letters are represented by '\_'.\\
There are two ways you can answer. You can provide one letter guesses in the format of \boxonly\{TOE\}, or you can guess the entire word in the format of \boxonly\{Y\}.\\
If the given letter is in the word, it will be revealed in the grid.\\
If the given word is correct, you win.\\
As you play, the history of your choices will be appended below. Use the information to figure out the word and win.\\
Some rules:\\
1. You can only guess one letter/word at a time.\\
2. You have to win within 10 turns.\\
Here is the current state of the Hangman grid:
\begin{verbatim}
C00 C01 C02
 _   _   _
\end{verbatim}
Enter your guess.
\end{tcolorbox}

\begin{tcolorbox}[colback=gray!10,colframe=black,title={Prompt for Blackjack}]
You are an agent playing a simplified game of Blackjack against a dealer.\\

Game Objective: \\
Your goal is to choose actions that maximize your chance of winning against the dealer.

- The objective is to get a hand value as close to 21 as possible without exceeding 21.\\
- If your hand value exceeds 21, you bust and immediately lose.\\
- After you stand, the game ends and your hand is compared with the dealer's hand.\\

Card Values: \\
- Number cards (2–10): face value\\
- Face cards (J, Q, K): value 10\\
- Ace (A): value 1 or 11, chosen to give the highest possible hand value not exceeding 21 \\

Dealer Rules: \\
- The dealer has exactly two cards.\\
- One dealer card is visible and the other is hidden.\\
- The dealer does not draw any additional cards.\\
- The dealer's hand value is computed using the same Ace rule as the player.\\

Initial Observation:\\
At the start of the episode, you observe:

1. Dealer’s hand:\\
   - One visible card and one hidden card\\
   - Example: Dealer cards: [4, ?]\\

2. Your hand:\\
   - A list of cards currently held\\
   - Example: Your cards: [10, 6]\\

3. Deck:\\
   - A list of remaining cards indexed by position\\
   - Unknown card identities are hidden\\
   - Example: Deck: [0: ?, 1: ?, 2: ?, 3: ?, 4: ?, ...]\\

Available Actions: \\
At each step, you must choose exactly one of the following actions:\\
- Hit: draw a specific card from the deck
  hit \textless card\_index \textgreater

- Stand: stop drawing cards and end the game
  stand\\

Action Output Format: \\
You must output your action in exactly one line, using the following format:

- Hit example: \boxonly\{hit 9\}

- Stand example: \boxonly\{stand\}\\

Important constraints:\\
- Output only the boxed action.\\
- Do not include explanations, reasoning, or additional text.\\
- Do not output multiple actions.\\

Now Begin: \\
Given the current observation, decide your next action and output it in the required format.
\end{tcolorbox}

\begin{tcolorbox}[colback=gray!10,colframe=black,title={Prompt for RockPaperScissors}]
You are playing a multi-turn Rock-Paper-Scissors game against an adversary. \\
In each episode, at every turn, the adversary's action is sampled from a fixed (hidden) distribution determined by the seed. \\
Your objective is to choose the action that maximizes the probability of winning against this hidden distribution each turn. \\
Now let's start the game. Output your action within \boxonly\{...\}.
\end{tcolorbox}

\begin{tcolorbox}[colback=gray!10,colframe=black,title={Prompt for Wordle}]
You are playing Wordle.\\
You have to guess the secret 5-letter word within 10 turns.\\
After you enter your guess, I will say mark your guess as follows:\\
\quad - G (green): correct letter in the correct position \\
\quad - Y (yellow): letter exists in the word but in the wrong position \\
\quad - X (wrong): letter is not in the word \\
After thinking, format your final answer inside \boxonly\{...\}, for example, CRANE. \\
As you play, the history of your guesses will be appended below. Use the information to complete the game before you run out of guesses.
\end{tcolorbox}

\begin{tcolorbox}[colback=gray!10,colframe=black,title={Prompt for Mastermind}]
You are playing Mastermind. \\
You have to guess the secret 3 digit code within 3 turns. \\
The code consists of digits from 1 to 6 (inclusive).\\
Duplicate numbers are 'not allowed'.\\
After you enter your guess, I will say mark your guess with black and white pegs, where a black peg indicates a correct digit in the correct position, while a white peg indicates a correct digit in the wrong position.\\
After thinking, format your final answer inside \boxonly\{...\}, for example, \boxonly\{1 4 6\}.\\
As you play, the history of your guesses will be appended below. Use the information to complete the game before you run out of guesses.\\
Enter your first guess to start the game.
\end{tcolorbox}

\begin{tcolorbox}[colback=gray!10,colframe=black,title={Prompt for Maze}]
You are a maze-solving agent. Your goal is to navigate from the START position to the GOAL position in the fewest turns possible. You are at the START position. \{observation\} \\
Output your next move from up/down/left/right within \boxonly\{\}.
\end{tcolorbox}

\section{Task Rules}\label{app:tasks}
We use seven games for either training or evaluation: Minesweeper, Hangman, Rock–Paper–Scissors, Wordle, Blackjack, Mastermind, and Maze. The rules of each game are summarized below.

\subsection{Minesweeper}
Minesweeper is a grid-based puzzle game in which some cells contain hidden mines. Revealing a safe cell displays the number of mines in its eight neighboring cells. If a revealed safe cell has zero neighboring mines, all adjacent cells are automatically revealed, potentially triggering further expansions. The objective is to reveal all non-mine cells without revealing any mine; players may place flags to mark suspected mine locations. The game ends either when a mine is revealed (failure) or when all safe cells have been successfully uncovered (success).

\subsection{Hangman}
Hangman is a word-guessing game in which a target word is selected and initially concealed, with each letter represented by a placeholder. On each turn, the player guesses a single letter. If the guessed letter appears in the word, all corresponding positions are revealed; otherwise, the number of remaining attempts is reduced by one. The objective is to fully reveal the target word before the attempt limit is exhausted.

\paragraph{Example}
Suppose the hidden word is \texttt{APPLE}. A guess of \texttt{A} reveals \texttt{A \_ \_ \_ \_} .  
A subsequent guess of \texttt{P} reveals \texttt{A P P \_ \_} .  
An incorrect guess, such as \texttt{Z}, does not reveal any letters and consumes one attempt.  
The game continues until the word is fully revealed or the attempt limit is reached.

\subsection{Wordle}
Wordle is a word deduction game in which a target word is selected and must be identified within a fixed number of attempts. On each turn, the player submits a word guess and receives feedback for each letter, indicating whether it is correct and in the correct position, correct but in an incorrect position, or absent from the target word. The objective is to deduce the target word before the attempt limit is reached.

\paragraph{Example}
Suppose the hidden word is \texttt{PLACE}.  
A guess of \texttt{ALIEN} receives feedback indicating that \texttt{L} is correct and in the correct positions, \texttt{A} and \texttt{E} are correct but in the wrong position, and \texttt{I} and \texttt{N} are absent from the word.

\subsection{Mastermind}
Mastermind is a code-breaking game in which a hidden sequence of numbers is randomly initialized. On each turn, the player submits a complete sequence as a guess and receives structured feedback. The number of black balls indicates how many numbers are correct and in the correct positions, while the number of white balls indicates how many numbers are correct but placed in incorrect positions. The objective is to identify the hidden sequence within a fixed maximum number of attempts.

\paragraph{Example}
Suppose the hidden code is (1, 2, 3).  
A guess of (1, 3, 4) receives feedback of one black ball (since 1 is in the correct position) and a white ball (since 3 is correct but not in the correct position).

\subsection{Maze}
Maze is a randomly initialized grid-world environment consisting of a starting position, walls, and a goal. Each instance is constructed such that the shortest path from the start to the goal lies within a pre-specified length range. At each step, the player observes only the four adjacent grid cells and selects an action to move within the maze. The episode is successful if the player reaches the goal within a fixed maximum number of steps; otherwise, it is considered a failure.

\subsection{Rock-Paper-Scissors}
Rock–Paper–Scissors is a repeated game in which the player competes against an adversary with a fixed stochastic policy. An episode consists of multiple rounds. In each round, both the player and the adversary simultaneously choose one of three actions: rock, paper, or scissors. At the end of the episode, the player is declared the winner if they win more than half of the rounds; otherwise, the episode is considered a failure.

\subsection{Blackjack}
Blackjack is a traditional card game. We consider a simplified setting in which the dealer is dealt two cards, one hidden and one revealed. The player also starts with two cards and may repeatedly choose whether to draw an additional card from the deck or to stop. Once the player stops drawing, the sum of the player’s cards is compared with the dealer’s. The player wins if their sum is greater than the dealer’s without exceeding 21; otherwise, the player loses. If the player’s card sum exceeds 21 at any point, the episode immediately ends in a loss.

\section{Oracle design}\label{app:oracle}
For the maze task, we implement a finite-horizon value-iteration oracle that plans on an incrementally constructed belief map aggregated from all previous observations. 
At each step, the oracle performs Bellman backups with memorization over the planning horizon, conditioned on the agent’s current position and accumulated observation history. 
The reward function assigns $100$ to wall collisions, $-0.1$ to revisiting known path cells, $100$ to reaching the goal, and $1$ to entering previously unvisited cells. This reward shaping incentivizes exploration for shortest-path discovery.

For the Mastermind task, we formulate the problem as a finite-horizon dynamic program over belief states that compactly encode the entire action–observation history, and obtain the optimal oracle via exhaustive enumeration of all possible guesses and feedback outcomes.

\section{Validation Reward Curves on Unseen Tasks}
\label{app:validation-curves}

\begin{figure}[t]
    \centering
    \includegraphics[width=0.65\columnwidth]{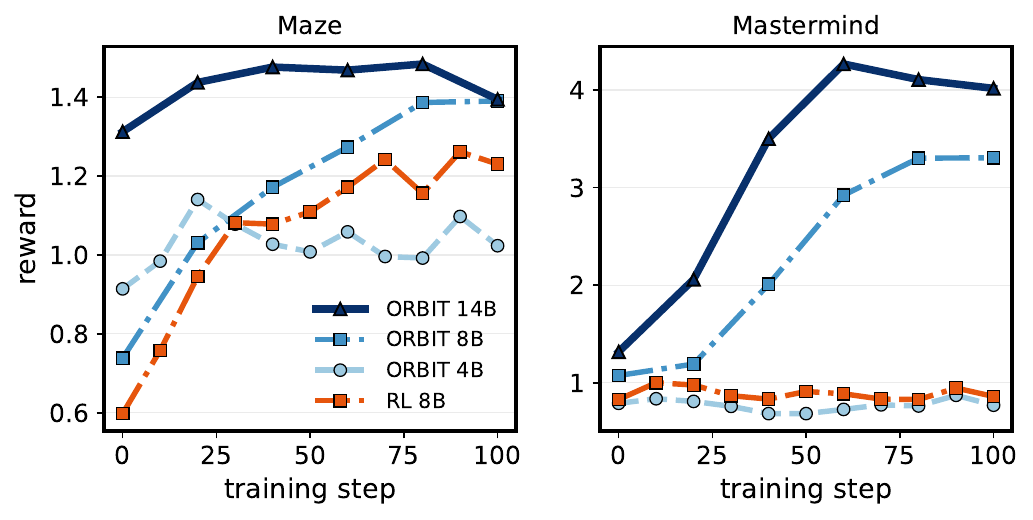}
    \caption{
    Validation reward versus training step on the unseen
    Maze (left) and Mastermind (right). Curves correspond to ORBIT
    models of different scales and a single-episode RL baseline.
    }
    \label{fig:validation-reward-curves}
\end{figure}

Figure~\ref{fig:validation-reward-curves} reports validation reward
(success count) as a function of training step on the unseen Maze and
Mastermind tasks.

\section{Additional Experimental Details}
All models are trained on a single node with 8 NVIDIA H100 GPUs. For OpenAI model evaluation, we use \texttt{gpt-5.2-2025-12-11} with high reasoning effort, and \texttt{gpt-4o-2024-08-06} with temperature \(0.6\) and top-\(p = 0.95\).

\end{document}